\documentclass{article}

\usepackage{arxiv}

\usepackage[utf8]{inputenc} 
\usepackage[T1]{fontenc}    
\usepackage{hyperref}       
\usepackage{url}            
\usepackage{booktabs}       
\usepackage{amsfonts}       
\usepackage{nicefrac}       
\usepackage{microtype}      
\usepackage{lipsum}		
\usepackage{graphicx}

\usepackage[numbers]{natbib}
\usepackage{doi}
\usepackage{amsmath, amssymb}
\usepackage{mathrsfs}
\usepackage{bm}
\usepackage[table]{xcolor}
\usepackage{appendix}

\usepackage[utf8]{inputenc}
\usepackage{amsmath,amssymb}
\usepackage{mathrsfs}
\usepackage{bm}
\usepackage{graphicx}
\usepackage[table]{xcolor}
\usepackage{booktabs}
\usepackage{graphicx}
\usepackage{multirow}
\usepackage{booktabs}
\usepackage{hyperref}
\usepackage{setspace}
\usepackage{algorithm}
\usepackage{algorithmic}
\usepackage{caption}
\usepackage{subcaption}
\usepackage{arxiv}
\usepackage[nohead, nomarginpar, margin=0.75in, foot=.25in]{}

\title{Tacit algorithmic collusion in deep reinforcement learning guided price competition: A study using EV charge pricing game}
\date{} 					

\author{ {\hspace{1mm}Diwas Paudel and Tapas K. Das}\\
        Department of Industrial and Management Systems Engineering, \\
	University of South Florida,\\
	Tampa, FL  \\
	\texttt{diwaspaudel@usf.edu,  das@usf.edu} \\
}

\begin{document}
\maketitle

\begin{abstract}
Players in pricing games with complex structures are increasingly adopting artificial intelligence (AI) aided learning algorithms to make pricing decisions for maximizing profits. This is raising concern for the antitrust agencies as the practice of using AI may promote tacit algorithmic collusion among otherwise independent players. Recent studies of games in canonical forms have shown contrasting claims ranging from none to a high level of tacit collusion among AI-guided players. In this paper, we examine the concern for tacit collusion by considering a practical game where EV charging hubs compete by dynamically varying their prices. Such a game is likely to be commonplace in the near future as EV adoption grows in all sectors of transportation. The hubs source power from the day-ahead (DA) and real-time (RT) electricity markets as well as from in-house battery storage systems. Their goal is to maximize profits via pricing and efficiently managing the cost of power usage. To aid our examination, we develop a two-step data-driven methodology. The first step obtains the DA commitment by solving a stochastic model. The second step generates the pricing strategies by solving a competitive Markov decision process model using a multi-agent deep reinforcement learning (MADRL) framework. We evaluate the resulting pricing strategies using an index for the level of tacit algorithmic collusion. An index value of zero indicates no collusion (perfect competition) and one indicates full collusion (monopolistic behavior). Results from our numerical case study yield collusion index values between 0.14 and 0.45, suggesting a low to moderate level of collusion.

\end{abstract}


\section{Introduction}

Dynamic pricing decisions by multiple competing players using reinforcement learning (RL)/deep reinforcement learning (DRL) methods have been presented in the recent economics literature. This body of literature considers multiplayer pricing games, mostly in canonical forms, and investigates the possibility of tacit algorithmic collusion among competing players that are guided by learning algorithms trained to maximize profit. Tacit algorithmic collusion refers to a behavior where players make decisions guided by algorithms that implicitly coordinate their actions to increase individual benefits without explicitly communicating or reaching an agreement. The algorithm-guided players signal their intentions through actions, which all players take note of in choosing their future actions.  The repetition of this learning process in RL/DRL algorithms is claimed to result in some form of tacit collusion, yielding profits higher than in a competitive market. Recent studies using Q-learning have examined Bertrand games (pricing games with unlimited supply) and indicated the presence of tacit algorithmic collusion. In the work presented in \cite{calvano2020artificial}, it is shown that the two competing firms using Q-learning algorithms can develop tacit collusion and charge prices that are significantly higher than the equilibrium prices. Similarly, the authors in \cite{mellgren2020tacit} corroborate the findings of \cite{calvano2020artificial} for competing firms that are guided by the DQN algorithm. However, the authors in \cite{den2022artificial} and \cite{zhang2023pricing} refute the above claims of strong evidence of tacit algorithmic collusion using their solutions obtained by Q-learning and DQN algorithms, respectively. Hence, there is a lack of consensus among the researchers on the likelihood of tacit algorithmic collusion among reinforcement learning-guided agents in pricing games. This paper aims to develop results to further our understanding of the tacit algorithmic collusion using a case of a pricing game among a set of competing EV charging hubs.

Fast-charging electric vehicle (EV) hubs will soon begin to replace gasoline refueling stations at street corners of cities across the world. These fast-charging hubs (hereafter referred to as hubs) will be central to the newly built infrastructure supporting the electrification of transportation systems  \cite{paudel2022infrastructure}\cite{alterenative_fuel}. We envision that the growing number of hubs will have the following characteristics. Multiple hubs serving an area will compete by dynamically adjusting their charging prices with the goal to maximize their profits. The hubs will coordinate with the power network for delivery and prices of electricity and with the transportation network for the EV charging demand. The hubs will use two primary sources for procuring electric power: the day-ahead (DA) market through a binding commitment for hourly quantities, and the real-time (RT) market for any additional power needs \cite{paudel2023distributionally}. A secondary source of electric power for the hub could be an in-house battery storage system (BSS). The hubs will use the BSS for arbitrage by effectively storing grid power and discharging it when profitable \cite{yan2018optimized}\cite{paudel2023deep}. The  EV charging demand arrival process will be subjected to several sources of randomness including traffic variations at different times of the day, battery sizes of the arriving EVs, and the charging preferences (price sensitivity) of the EV owners. Effective selection of dynamic pricing strategies will be a crucial profit-maximizing task for the competing hubs. Hence, a dynamic pricing model for the hubs must simultaneously consider variations in the DA and RT prices, randomness in demand arrival, the price and availability of BSS power, and the price sensitivity of EV owners.

At the core of the competition among the hubs is a Bertrand-Edgeworth model of price-setting oligopoly with homogeneous products (charging power) and limited supply capacities (fixed number of charging stations). For this class of problems, Edgeworth pointed out the non-existence of pure strategy price equilibrium and that the prices cycle within some bounds \cite{hicks1935annual}.  Subsequent literature claims the existence of mixed-strategy price equilibrium under specific cost, capacity, and price assumptions \cite{dixon1984existence}. However, in games like the hub-pricing game, these simplified assumptions do not hold. Moreover, the learning algorithm-guided pricing solutions may not follow the above claims. In this work, we develop a methodology for obtaining deep reinforcement learning (DRL) solutions for Bertrand-Edgeworth games using the hub pricing competition. We analyze the nature of the resulting pricing strategies and investigate the presence of tacit algorithmic collusion. 

Dynamic pricing models presented in the literature have used reinforcement learning (RL) and deep reinforcement learning (DRL) algorithms for diverse application areas including perishable products \cite{burman2021deep}, online marketplaces \cite{kastius2021dynamic}, and EV charging hubs \cite{aljafari2023electric}. A Deep Q-Network (DQN) algorithm is used in \cite{burman2021deep} to obtain revenue-maximizing prices for a single player dealing with perishable products. The DQN policy is shown to yield higher revenue compared to the myopically optimized prices or fixed prices.  A dynamic pricing model using DRL methods for an online marketplace \cite{kastius2021dynamic} finds that for a two-player game, the strategies derived by DQN and Soft Actor-Critic (SAC) algorithms outperform other heuristic methods. A DRL-aided dynamic pricing strategy for EV charging/discharging is developed in \cite{aljafari2023electric}, which guides EVs to discharge power (vehicle to grid) at peak price hours and charge (grid to vehicle) during off-peak hours. In this multi-agent DRL approach, the first agent aims at minimizing the cost of charging the EVs while the second agent aims at maximizing the revenue from discharging the EVs. The paper in \cite{fang2020dynamic} presents a dynamic adjustment of retail EV charging price to maximize the profit for a fast charging hub (without any competition). The paper uses both RL and DRL approaches to derive pricing decisions considering price-elastic EV charging demand. Similar works using DRL methods for dynamic pricing can be found in \cite{narayan2022dynamic}, \cite{abdalrahman2020dynamic}, and \cite{liu2021dynamic}, to cite a few. These papers study dynamic pricing decisions for single hubs considering price-elastic charging demand. To our knowledge, a DRL approach for making dynamic pricing decisions for multiple competing fast-charging hubs that source electricity from the DA and RT markets as well as the battery storage system has not yet been examined in the open literature. 

The following are the key contributions of this paper. 
\begin{itemize}
    \item We examine the presence of tacit algorithmic collusion in practical supply-constrained pricing games with continuous system state and action spaces by developing a deep reinforcement learning-guided dynamic pricing methodology. 

    \item The methodology is demonstrated using a price competition among the players (EV charging hubs) that procure goods (power) from multiple sources with random price variations (day-ahead and real-time electricity markets and battery storage system) and meet demands of price-sensitive consumers (EV owners) under supply constraint (a finite number of charging stations). 
    
     \item An index is used to quantify the level of tacit algorithmic collusion among the players.

     \item We study the impact of heterogeneity of choices by the players for combinations of different DRL algorithms and neural network architectures on the pricing strategies and resulting collusion levels. 
      
\end{itemize}

The rest of this paper is organized as follows. We describe the EV charge pricing game in Sections \ref{sec: ev charge game}  and provide the system model in Section \ref{sec: system model}. The details of our solution methodology for the system model are provided in Section \ref{sec: solution approach} followed by a numerical case study in Section \ref{sec: Numerical case study} and sensitivity analysis in Section \ref{sec: sensitivity analysis}. Finally, the concluding remarks are provided in Section \ref{sec: conclusions}.

\section{EV charge pricing game}\label{sec: ev charge game}
\begin{figure}[htp]
    \centering
    \includegraphics[width=0.75\linewidth]{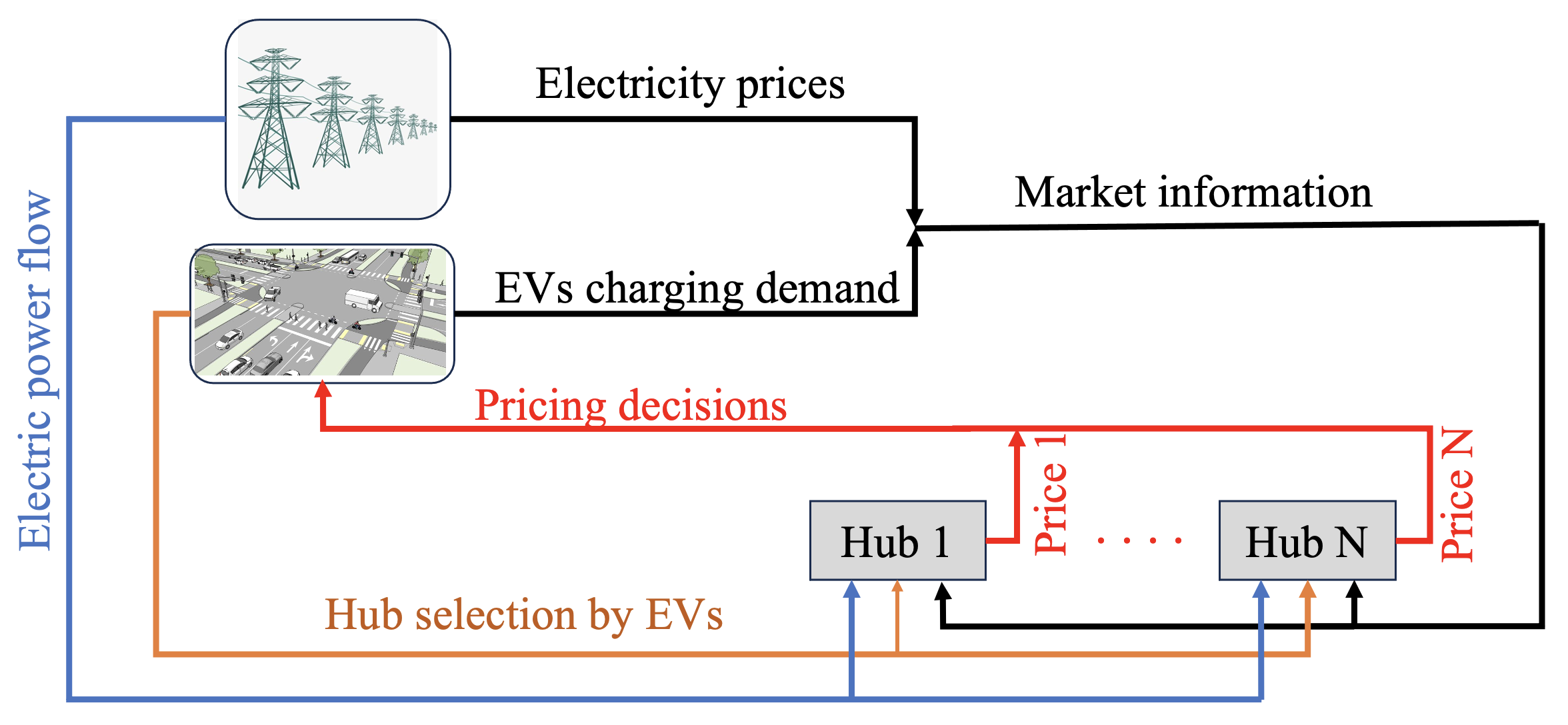}
    \caption{Schematic of EV charge pricing game}
    \label{fig: problem_description}
\end{figure}

The growing adoption of EVs presents a significant new business opportunity for fast-charging hubs. A number of such hubs, often with a large number of DC fast charging stations, located in close vicinity will compete by dynamically varying their prices to attract charge-seeking EVs in the area. Price-sensitive EV owners select the cheapest hub with availability. The hubs use electric power from the grid and in-house battery storage systems to meet the EV charging demand. The hubs aim to maximize their individual profits through pricing. Henceforth we refer to this price competition as a hub pricing game. The mechanism of the hub pricing game, depicted in Figure \ref{fig: problem_description}, is discussed next. 

For each time period, the hubs gather current market information on electricity prices and EV charging demand. This together with the state of their in-house battery storage systems (i.e., quantity and price of the stored power), the hubs make their pricing decisions simultaneously. Since the electricity prices can vary every hour for DA and in much shorter intervals for RT, hubs can revisit their pricing decisions as frequently as every hour or less. However, synchronized pricing actions by the hubs is an assumption made for model tractability. The EV owners respond to the hub prices by either selecting the cheapest available hub or deciding not to charge (balking). The balking decision by an EV owner may be influenced by the perception of how high the current lowest available price is and the EV's state of charge (SOC). The response of EV owners determines the number of EVs in each hub. The sizes and SOC of the EV batteries together with the owners' charging preferences determine the charging power demand in each hub. Each hub strives to meet charging demand by minimizing the cost of procuring power from the grid and the battery storage system. Note that the power grid supplies electricity using two different market mechanisms, namely the day-ahead (DA) market, where the hubs must commit to their hourly purchase levels ahead of the next day, and the real-time (RT) market for additional power needs. The hubs use their in-house battery storage systems for arbitrage by storing and discharging electricity when profitable. The hubs can store electricity in BSS from both DA and RT markets while BSS is discharged only for EV charging.  

\section{System model}\label{sec: system model}
\begin{figure}[htp]
    \centering
    \includegraphics[width=1\linewidth]{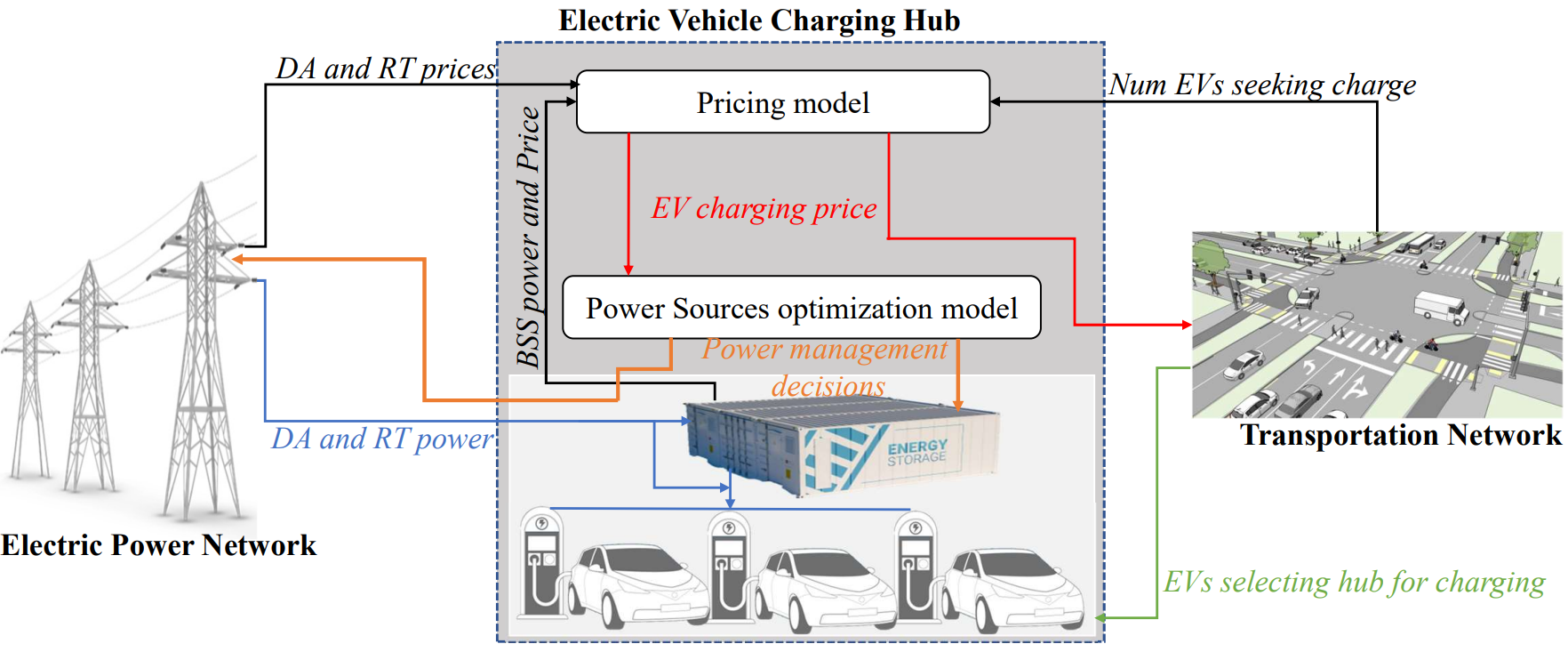}
    \caption{A hub's interaction with transportation and power networks}
    \label{fig:EVCH structure}
\end{figure}

 A simulation model emulates the operation of three interacting system components: the transportation network, the electric power network, and the EV charging hubs (see Figure \ref{fig:EVCH structure}). In what follows, we describe the models representing each component. 

\subsection{Transportation network}\label{trans_net}
The total EV charging demand served by the competing hubs in each time period of the day is modeled as follows. Using available traffic data, we determine the aggregate number of cars $N_t$ traveling per unit time $t$ (say, an hour) in the road network. Based on the prevailing EV penetration percentage $\beta$, the total number of EVs on the roads is calculated as $\beta N_t$. We assume that a certain percentage ($\alpha$) of the EVs on the road will use public fast-charging hubs. Hence, the total number of EVs that might seek to charge at any time period is $\alpha\beta N_t$. Since the traffic flow is highly stochastic, we use $\alpha\beta N_t$ as the rate parameter for a Poisson distribution to generate the number of EVs that might use the fast-charging hubs that are considered $({\hat N}_t)$. Using the probability $(p_t)$ of an EV actually seeking to charge at any time period $t$, the actual number of charge-seeking EVs $n_t$ is then obtained from a binomial probability distribution with parameters $({\hat N}_t, p_t)$. The EVs respond to the pricing decisions by the hubs as follows. 

Let $\gamma$ percent of the EV owners are price sensitive and they select the lowest-priced charging hub with available capacity. However, if the price difference between two hubs is less than $\delta$ percent, then the EV owners are assumed to select either with equal probability. The EV owners balk (i.e., decide not to charge) with probability $p_k$ if the lowest price of the currently available hubs is significantly higher (i.e. price difference greater than $\delta\%$) than the cheapest hub at that time which is already full. Price-insensitive EV owners, $(1-\gamma)\%$, select any of the available hubs with equal probability. 

\subsection{Electric power network}
The charging hubs interact with the electric power network by participating in day-ahead and real-time markets through the wholesalers. It is considered that hubs' actions in the power markets do not directly influence the DA and RT prices, and hence the hubs are price takers. The hubs commit their hourly quantities ($da_t^{com}$) in the DA market. After the market clears, the hubs receive hourly DA prices ($p_t^{da}$) for the following day. In each time-period, the hubs acquire any additional power needs from the RT market and battery storage. It is assumed that the hubs can sell back any unused DA commitment to the RT market, for which they are paid lower of the DA and RT prices. 

\subsection{EV charging hubs}
The hubs determine their day-ahead commitment by solving a scenario-based stochastic optimization model. For each time period of the day, given the DA commitment, BSS state, RT prices, and the total charging demand, each hub makes its pricing decision using a deep reinforcement learning algorithm. The pricing decisions together with the price response by the EVs determine the distribution of the total charging demand among the hubs. Finally, each hub solves a mixed integer power management model to meet its EV charging demand. The models for day-ahead (DA) commitment, competitive dynamic pricing, and hub power management are presented next. 

\subsubsection{Day-ahead commitment}\label{section: DA commitment}
The hourly DA commitment considers anticipated DA and RT prices and EV charging demand. A stochastic optimization model yields the DA commitment by considering a number of representative scenarios for DA and RT prices and aggregated EV charging demand.
Since the DA commitment is made before the pricing decisions, we adopt a conservative approach of using the minimum of the DA and RT prices as the EV charging price in the DA model. Likewise, as the actual EV charging demand for a hub is not determined until it makes the pricing decisions, the DA commitment model assumes that the total charging demand is distributed among the hubs proportional to their sizes. The mathematical formulation of the DA commitment model is provided in Appendix \ref{Appendix: DA commitment model}. 

\subsubsection{Competitive dynamic pricing}\label{section: CMDP}

At the start of any time period $t$, based on the current system state, the hubs simultaneously make their pricing decisions and the system evolves to the next state. 

Let $\mathcal{S}_t$ denote the component of the system state that is observable to all the hubs and is given by a vector comprising the total number of EVs seeking to charge $(n_t)$, DA price $(p_t^{da})$, and RT price $(p_t^{rt})$. Also let   $\widehat{\mathcal{S}}_t^i$ denote the system state component that is only visible to hub $i \in \mathcal{I}$ and is comprised of the hub's DA commitment $(da_t^{i, com})$, stored power in the BSS $(\phi_t^i)$, and the price of BSS power $(p_t^{i, \phi})$. Then $(\mathbf{S}, \mathbf{\widehat{S}})$ denote the system state process, where $\mathbf{S}=\{S_t:\forall t\in \mathcal{T}\}$ represents the fully observable component of the system state process, and $\mathbf{\widehat{S}}=\{\widehat{S}_t^i: \forall i \in \mathcal{I}, \forall t\in \mathcal{T}\}$ represents the partially observable component of the system state.  
Let $A_t^i$ denote the pricing decision by hub $i$ in time $t$ and $R_{t}^i$ denote the reward for the decision $A_t^{i}$. Then $\mathbf{A} = \{\mathcal{A}_t^i: \forall i\in \mathcal{I}, \forall t \in \mathcal{T}\}$ represents the decision process and $\mathbf{R} = \{\mathcal{R}_t^i: \forall i\in \mathcal{I}, \forall t \in \mathcal{T}\}$ represents the reward process. Since $(\mathbf{S}, \mathbf{\widehat{S}})$ is a Markov process, the process denoted by $\big(\mathbf{S}, \mathbf{\widehat{S}, A, R}\big) = \big\{(\mathcal{S}_t,\widehat{\mathcal{S}}_t^i, \mathcal{A}_t^i, \mathcal{R}_t^i): \forall i \in \mathcal{I}, \forall t \in T \big\}$  is a competitive Markov decision process (CMDP). The solution of this CMDP yields the pricing strategies for the competing hubs.

\subsubsection{Hub power management}
For each time period, once the charging demands for the hubs are determined, each hub engages in the management of its power sources to minimize the cost of power usage and thereby maximize gross profit. For this, each hub solves a mixed integer linear programming (MILP) model (see Appendix \ref{Appendix: Power mgmt}). The decision variables in the MILP model are DA power used for EV charging ($da_t^{i,ev}$), DA power used for BSS charging ($da_t^{i,bss}$), DA power sold back to the RT market ($da_t^{i,rt}$), RT power used for EV charging ($rt_t^{i,ev}$), and BSS power used for EV charging ($bss_t^{i,ev}$).     

\section{Solution methodology for the system model}\label{sec: solution approach}
We develop a two-step methodology for solving the components of the system model described above. The solution yields the dynamic pricing strategies which are then used to calculate the level of tacit algorithmic collusion among the hubs. The first step of the methodology involves solving the DA commitment model using any optimization model solver. In the second step, we solve the CMDP model using a multi-agent deep reinforcement learning (MADRL) framework, where each hub adopts a DRL algorithm and an underlying neural network (NN) architecture to guide its pricing decisions. After the pricing decisions are made at the top of each time period and the charging demands are realized, each hub solves its power management model. 
The objective function values (gross profit) from the power management model serve as the rewards for the hubs' pricing decisions for the time period. The system state transitions to a new state at the end of the current time period and this transition is orchestrated via the system simulation model. The training of the DRL algorithms includes alternating execution and evaluation steps continued for a large number of episodes (days). During the execution step, the pricing decisions are obtained from the neural networks associated with each hub and are implemented in the simulation model. In the evaluation step, the neural network weights for each hub are updated using the gradients of the loss calculated from the rewards. The trained neural network weights represent the dynamic pricing strategies of the hubs. Further details (pseudo code) of our two-step solution methodology are presented in Algorithm \ref{algo: two-step methodology}. 

\begin{algorithm}[htp]
    \caption{Solution methodology for the system model}\label{algo: two-step methodology}
    \begin{algorithmic}
        \STATE \textbf{STEP 1: }Generate day-ahead commitment quantities
        \FOR{each hub $i \in I$}
            \STATE Generate a large number of scenarios of DA prices, RT prices, and expected EV charging demand. 
            \STATE Use a scenario reduction technique to select a set of most likely representative scenarios.
            \STATE Using these scenarios, solve the DA commitment model to obtain hourly quantities.
        \ENDFOR
        \STATE \textbf{STEP 2:} Solve the CMDP using MADRL approach
        \STATE  Initialize the execution and target neural networks' weights for each hub $i \in \mathcal{I}$
        \STATE Initialize the simulation environment 
        \STATE Initialize replay memory $\mathcal{D}_i, \forall i \in \mathcal{I}$
        \FOR{each training episode (day)}
            \FOR{each decision period $t \in T$}
                \FOR{each hub $i \in \mathcal{I}$}
                    \STATE Determine system state ($\mathcal{S}_t, \widehat{\mathcal{S}}_t^i$)
                    \STATE Pass the system state through the execution neural network to obtain the pricing decision ($p_t^{i,ev})$ 
                \ENDFOR
                \STATE Make the pricing decisions ($p_t^{i,ev}, \forall i \in \mathcal{I})$ available to the EV owners for hub selection
                
                \FOR{each hub $i \in \mathcal{I}$}
                    \STATE Determine aggregated charging demand 
                    \STATE Solve the power management model and observe the reward $R_t^i$
                    \STATE Observe the next state and store the transition $\big\{ (\mathcal{S}_t, \widehat{\mathcal{S}}_t^i), p_t^{i,ev}, R_t^i, (\mathcal{S}_{t+1}, \widehat{\mathcal{S}}_{t+1}^i)\big\}$ in $\mathcal{D}_i$
                \ENDFOR
            \ENDFOR
        \ENDFOR
        \FOR{each evaluation step}
            \FOR{each hub $i \in \mathcal{I}$}
                \STATE Sample a batch of transitions from $\mathcal{D}_i$
                \STATE Calculate the loss and update the execution networks 
                \STATE Update the target networks
            \ENDFOR
        \ENDFOR
    \end{algorithmic}
\end{algorithm}

\newpage

\section{Numerical case study}\label{sec: Numerical case study}

\begin{figure}[htp]
    \centering
    \includegraphics[width=0.5\linewidth]{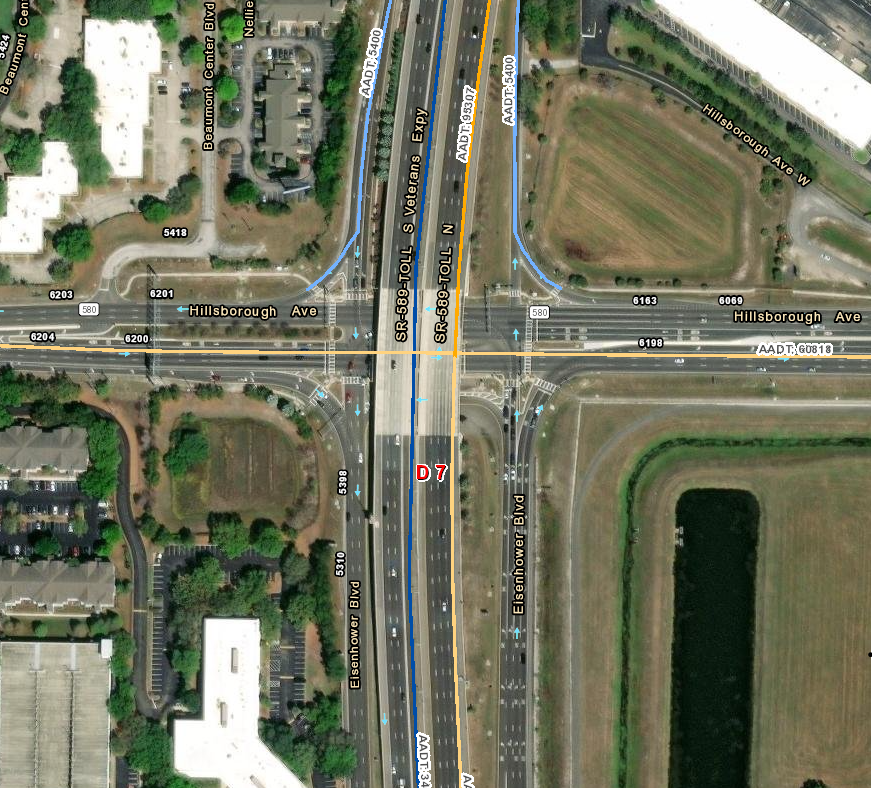}
    \caption{Road intersection in Tampa, Florida, USA (between Hillsborough Avenue and Veterans expressway) considered to generate the EV traffic data. The competing hubs are assumed to be located at this intersection.}
    \label{fig: intersection}
\end{figure}

\begin{figure}[htp]
    \centering
    \includegraphics[width=0.65\linewidth]{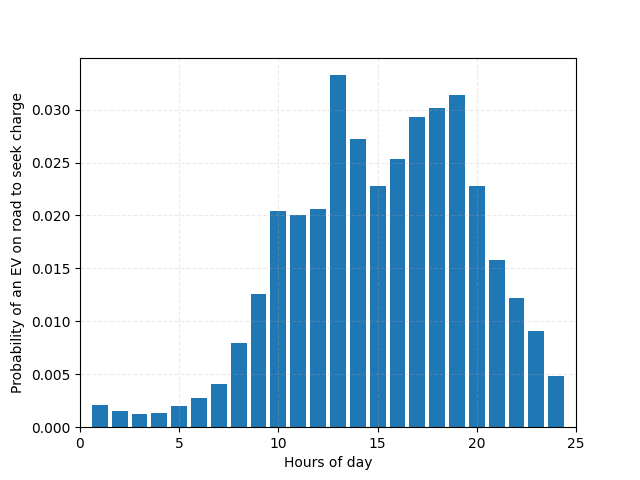}
    \caption{Probability of an EV on the road to seek charge over the hours of a day \cite{deng2018demand}.}
    \label{fig: probab of ev to seek charge}
\end{figure}

\begin{figure}[htp]
    \centering
    \includegraphics[width=1\linewidth]{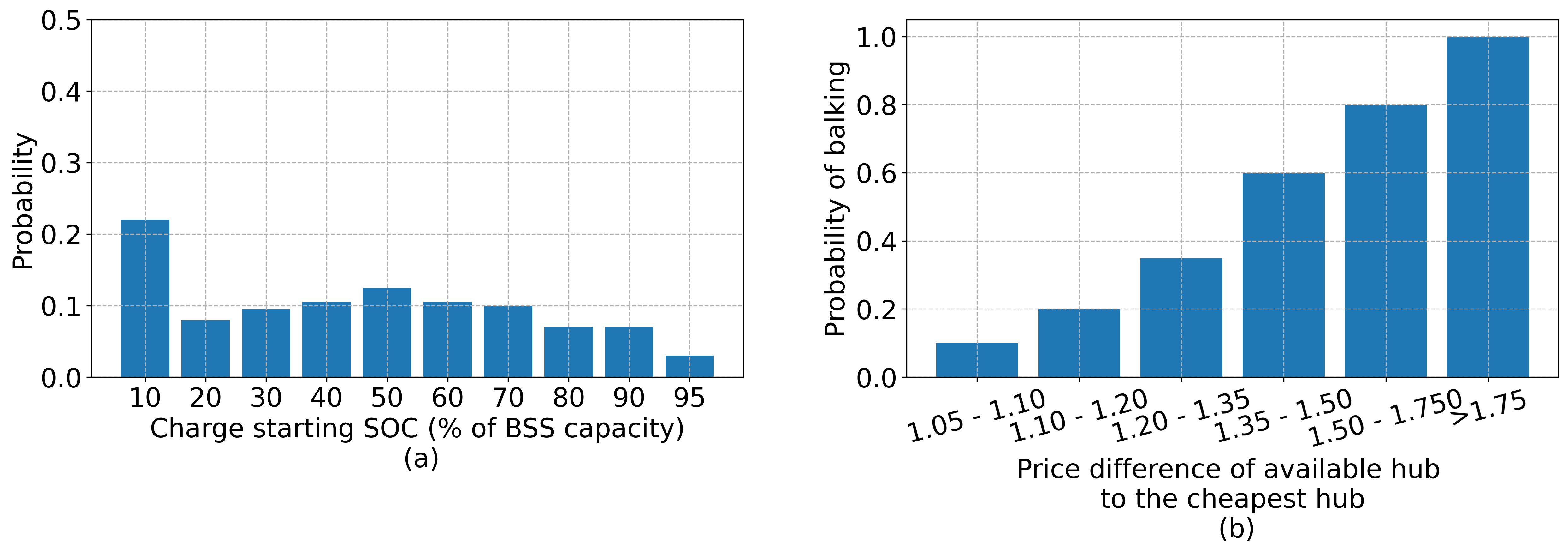}
    \caption{(a) probability distribution for EV state of charge(SOC) in percentage at the start of charging \cite{Idaho}, and (b) probability of balking by an EV for various price ratios of the hub that is available for charging to the cheapest hub.}
    \label{battery characteristics}
\end{figure}


To examine the pricing strategies from our methodology, we develop a sample numerical case study with two identical hubs in price competition. Each hub has 150 DC fast-charging stations with a maximum charge rating of 100 kWh and a battery storage system (BSS) with 4000 kW storage capacity. The BSS is considered to charge/discharge up to 2000 kW per hour and maintains a minimum charge of 500 kW; similar considerations are found in  \cite{funke2020fast}, \cite{hussain2020optimal}. The number of EVs seeking to charge in any time-period (say, an hour) is obtained as follows. Hourly traffic flow data $(N_t)$ are collected from the Florida Department of Transportation \cite{traffic_flow} for the road intersection where the hubs are assumed to be located (see Figure \ref{fig: intersection}). This is done for the seventh day of every month of the year 2020 and the hourly averages ($\overline{N_t}$) are calculated. It is assumed that 25\% of the traffic is EV \cite{McKinsey} ($\beta = 25\%$) and 42\% of those EVs use public fast-charging hubs ($\alpha = 42\%$) \cite{narayan2022dynamic}. Then the average values of the EVs that flow each hour through the intersection and use public charging hubs are obtained as $\alpha\beta\overline{N_t}$. We use these average values as the rate parameter for Poisson distributions to generate the number of EVs flowing through the intersection and depend on the use of public charging facilities ($\hat{N_t}$). Finally, we generate the actual number of EVs that will seek to charge in any given hour ($n_t$) using a Binomial distribution with parameters $(\hat{N_t}, p_t)$, where $p_t$ is the probability of an EV near the intersection receiving charge at time $t$. This probability is obtained from the EV charging behavior study conducted by Idaho National Lab \cite{deng2018demand}. The EV charging demand generation process is summarized in Algorithm \ref{algo: EV charging demand}. 

EVs are considered to have three different battery sizes, 50 kW, 75 kW, and 100 kW with probabilities of 0.3, 0.4, and 0.3, respectively. The charge that an EV seeks to receive varies between 10\% to 95\% of the battery size \cite{Idaho}. All the EV owners are considered to be price sensitive ($\gamma = 1$). It is assumed that the prices are bounded between the minimum of the DA and RT prices, and two times the minimum. EV owners are assumed to choose an available hub with either the minimum price or those within 5\% of the minimum. EV owners are assumed to balk (i.e., not charge), with a certain probability, if the available hub has a price significantly higher than the minimum price for that time-period. The balking probability increases with the price differential (see Figure \ref{battery characteristics}c). The hubs always fulfill the demand of the arriving EVs using power from three sources: DA commitment, RT market, and BSS. It is assumed that DA commitment cannot be used for arbitrage in the RT market. Only excess DA commitment after EV and BSS charging is sent back to the RT market, for which the hubs receive lower of the prevailing DA and RT prices. This ensures that DA commitment is used primarily for EV charging.  

\begin{algorithm}[htp]
 \begin{algorithmic}
    \STATE Obtain average traffic flow data in the intersection for each hour of a day $(\bar{N_t})$\;
    \begin{itemize}
        \item 25\% ($\beta$) of the traffic are considered to be EVs\;
        \item 42\% ($\alpha$) of EVs use public charging facilities
    \end{itemize}
    \STATE Calculate the average number of hourly EVs that use public charging facilities as $\alpha\beta \bar{N_1}, \alpha\beta \bar{N_2}, \dots, \alpha\beta \bar{N_{24}}$\; 
    \FOR{day in 365 days}
        \FOR{hour of the day}
            \STATE Calculate the actual number of EVs that will seek to charge $(n_t)$  as Binomial $(\alpha\beta \bar{N_t}, p_t)$, where $p_t$ is the probability of an EV deciding to charge in time $t$.
        \ENDFOR
        \ENDFOR
\end{algorithmic}
\caption{EV charging demand generation}
\label{algo: EV charging demand}
\end{algorithm}

\subsection{DRL algorithms and Neural  network architectures}
We implement two different DRL algorithms to learn the competing pricing strategies of the hubs. The algorithms are DQN \cite{mnih2013playing}, a deep variant of Q-learning, and SAC \cite{haarnoja2018soft}, a variant of the policy gradient algorithm. Existing literature on RL-based approaches for pricing games considers canonical formulations and solves them using Q-learning and DQN algorithms. Since our problem has continuous state space, we chose to implement DQN, for which however we had to discretize our continuous action space. For SAC implementation, we were able to consider continuous state and action spaces, which are inherent to our problem.  
As regards neural network architecture, we consider a feed-forward (FF) and a multi-head attention (MHA) network \cite{vaswani2017attention}. We consider two variants of DRL algorithms and two different neural network architectures to examine the impact of different algorithm-architecture combinations that may be adopted by independent competing hubs. We provide the overview of DRL algorithms in Appendix \ref{Appendix: DRL algo}, schematic representations of neural network architectures in Appendix \ref{Appendix: network}, and the hyperparameters in Appendix \ref{Appendix: Hyperparameters}.

\subsection{Train and test data sets}
For our numerical study, we obtain the required training and test data as follows. Hourly DA and RT prices for 365 days (Dec. 2021 till Nov. 2022) are collected from the Pensylvania-Jersey-Maryland (PJM) interconnection archive \cite{PJM}. EV charging demand scenarios are also generated for 365 days using the algorithm \ref{algo: EV charging demand}. From this combined data set comprising 365 scenarios,  we randomly select 8 scenarios from each of the four seasons of the year (a total of 32 scenarios) and set aside as the test data set. The remaining 333 scenarios are used for training of the DRL agents. 

\subsection{Implementation of methodology}
We first apply a scenario reduction technique \cite{green2014divide} on the training dataset to select a set of ten representative scenarios of DA/RT prices and EV charging demands with their respective probabilities. Considering these scenarios, we solve our stochastic DA commitment model using Gurobi 9.5.2. Since we consider two hubs to be identical in our problem, the same DA commitment is used by both in the pricing decision-making step of the methodology. The MADRL algorithms are implemented using Pytorch 2.0. The hyperparameter values associated with the neural network architectures and the DRL algorithms are provided in Table \ref{table: hyperparameters} (in Appendix \ref{Appendix: Hyperparameters}). All the computations are performed using an Intel i9-11900H@2.50GHz processor with 32 GB RAM and NVIDIA GeForce RTX 3080 GPU with 8 GB memory. Both the simulation environment and the algorithms are implemented using Python 3.8. 

\subsection{Results}

\begin{figure*}[htp]
    \centering
    \includegraphics[width=1\linewidth]{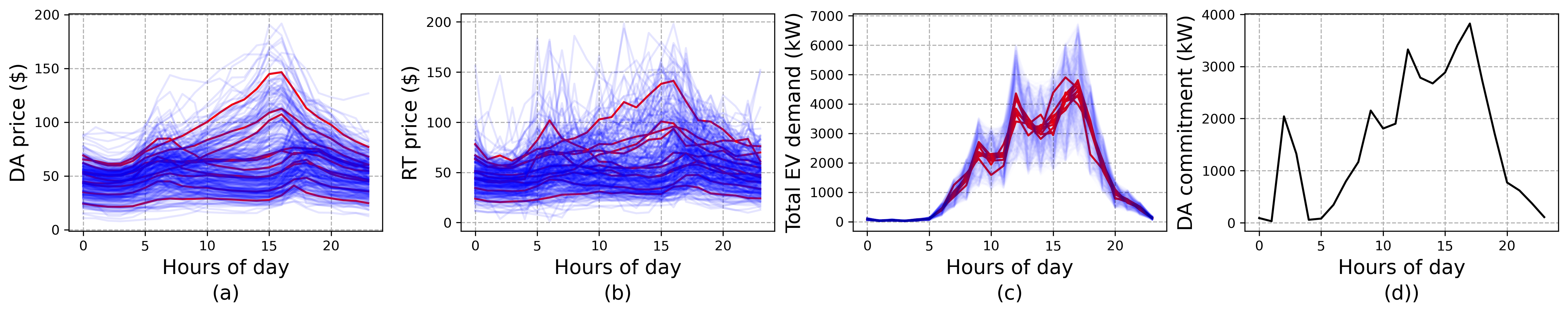}
    \caption{(a) Actual scenarios (blue lines) and reduced scenarios (red lines) of DA prices, (b) RT prices, (c) total EV charging demand for a hub (d) resulting DA commitment for each hub}
    \label{fig: da commitment}
\end{figure*}

\begin{figure}[htp]
    \centering
    \includegraphics[width=1\linewidth]{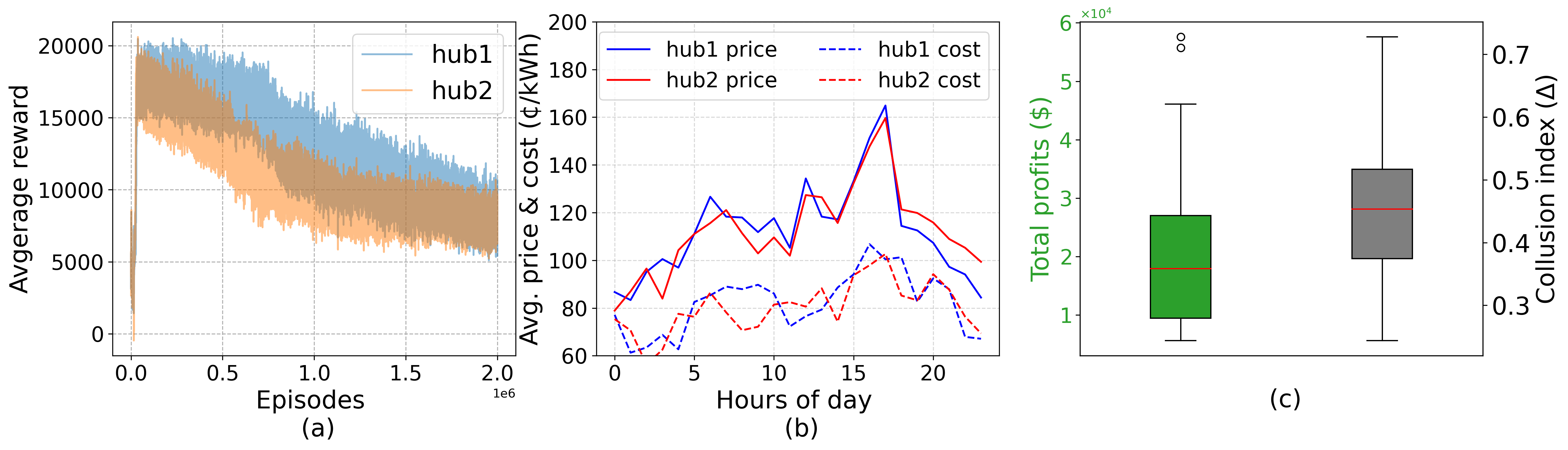}
    \caption{ (a) Learning curves, (b) average prices and costs, and (c) box plots for total profit and collusion index (when both hubs use SAC-FF algorithm-architecture combination)}
    \label{fig: reward_curves}
\end{figure}

Figure \ref{fig: da commitment} shows the outcomes of the DA commitment step of our methodology. Figure \ref{fig: da commitment}a, \ref{fig: da commitment}b, and \ref{fig: da commitment}c depict the selected scenarios (in red) from the historical data (in blue) for DA/RT prices and the EV charging demand. Figure \ref{fig: da commitment}d presents the hourly DA commitment values that maximize the expected profit for the hubs over the selected scenarios. It can be observed that the DA commitment follows a similar profile to that of the EV charging demand. This is expected, as the purpose of making a DA commitment is to hedge against real-time price volatility. We notice, however, that for the early hours of the day, the DA commitments are much higher than the EV charging demands. This excess power purchase is for the hubs to charge their BSS, as the DA power is generally cheaper in the early hours of the day. The DA commitment is used as input for the second step of our methodology, which is the real-time dynamic pricing model. The results from the dynamic pricing model are presented next. 

We first examine the results from the case where both the hubs use the SAC algorithm and a feed-forward (FF) neural network architecture. We chose this as our base algorithm-architecture combination since this can accommodate continuous state and action spaces inherent to our problem. Other algorithm-architecture combinations are examined through a sensitivity analysis presented later. Figure \ref{fig: reward_curves}a presents the learning curves for the training of our base combination, for which the stopping criterion was either no significant change noted in the average rewards or two million training episodes completed. The neural network weights at the end of the training experiment represent the dynamic pricing strategy for the hubs. These pricing strategies are implemented in the 32 test data scenarios. The resulting hourly prices averaged over all test scenarios are depicted in Figure \ref{fig: reward_curves}b, which also includes the average hourly cost (cents per kWh) incurred by the hubs in procuring the power needed for EV charging. This hourly cost of power is a combination of the prices of DA, RT, and BSS power based on their usage as suggested by the power management model (see Appendix \ref{Appendix: Power mgmt}). Note from the figure, that the average hourly prices for two identical competing hubs are similar with minor differences across the hours, which is expected. The hourly costs incurred by the two hubs are also similar. The prices appear to follow a similar pattern to that of the costs with an average markup of 1.38 overall hours of the day. This markup is indicative of potential collusion among the hubs. However, since the hubs learn pricing decisions through experience without any explicit information sharing, collusion in such situations is referred to as tacit algorithmic collusion \cite{schwalbe2018algorithms} \cite{ittoo2017algorithmic}. In what follows, we present an index, a measure that we use to quantify the level of tacit algorithmic collusion, and use it in our numerical case study. 

Using the definition presented in \cite{calvano2020artificial}, we construct our collusion index $(\Delta)$) as follows.
\begin{equation}
    \Delta = \frac{\overline{\pi} - \pi^C}{\pi^M - \pi^C},
\end{equation}
where, $\overline{\pi}$ is the total daily profit made by the hubs using their DRL-guided pricing strategies, $\pi^C$ is the total daily profit in pure competition when hubs select the prices same as the costs, and $\pi^M$ is the total daily profit when the hubs price with full collusion (pure monopoly). A value of $\Delta = 0$ corresponds to pure competition and $\Delta = 1$ corresponds to full collusion. Figure \ref{fig: reward_curves}c, provides the box plots for the total daily profit and the collusion index from the 32 test data scenarios. The median value of the collusion index is 0.45. This value is in between what has been described in the literature as a high level of collusion ($\Delta$ greater than 0.7 \cite{calvano2020artificial}) and almost no collusion ($\Delta$ less than or equal to 0.02 \cite{zhang2023pricing}). 

\begin{figure}
    \centering
\includegraphics[width=1\linewidth]{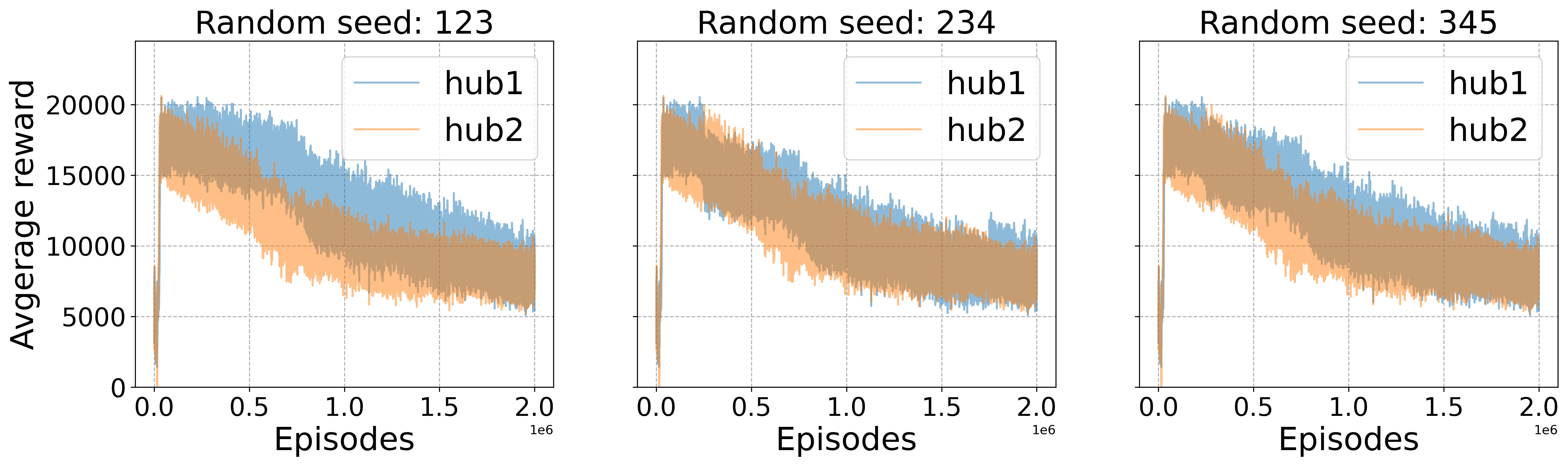}
    \caption{Learning curves under different random seeds}
    \label{fig: randomseed}
\end{figure}

We also implemented our SAC-FF experiment with two additional values of random seeds to examine the robustness of our numerical results. The learning curves for the new seeds have similar average reward values, as for the original seed, at 2 million training episodes (see Figure \ref{fig: randomseed}). We also found the average hourly prices from these new seeds to be close to those from the original seed.

\section{Sensitivity Analysis}\label{sec: sensitivity analysis}

\begin{figure*}[ht]
    \centering
    \includegraphics[width=1\linewidth]{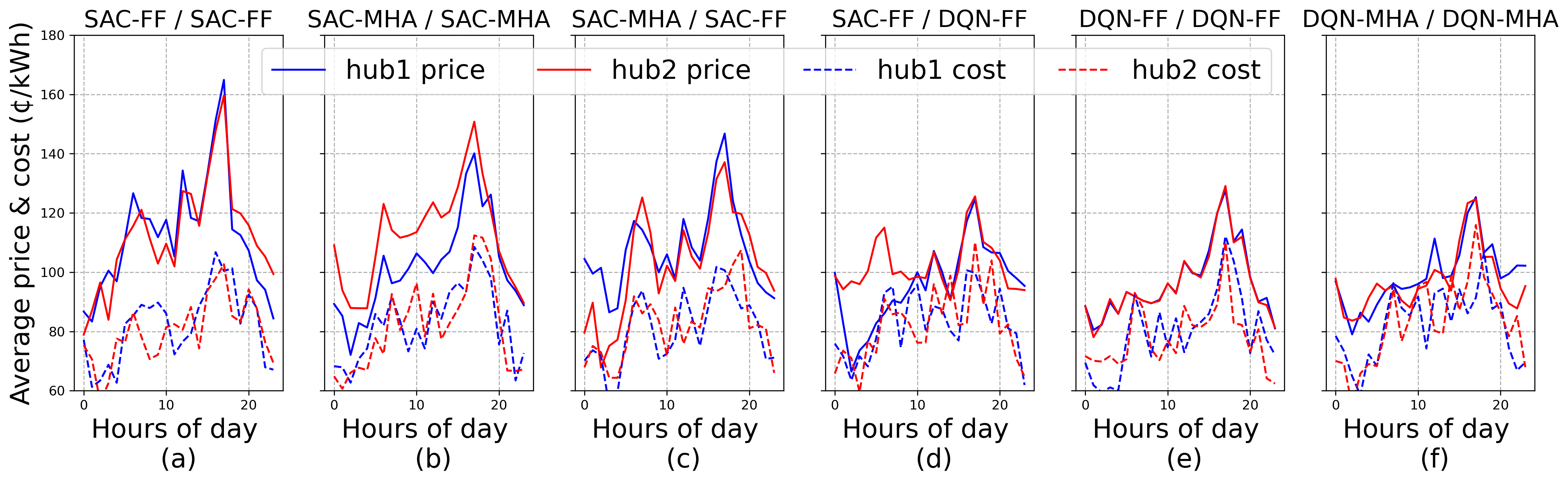}
    \caption{Hub pricing strategies and cost of power averaged for 32 test scenarios}
    \label{fig: price_curves}
\end{figure*}

\begin{figure*}[ht]
    \centering
\includegraphics[width=1\linewidth]{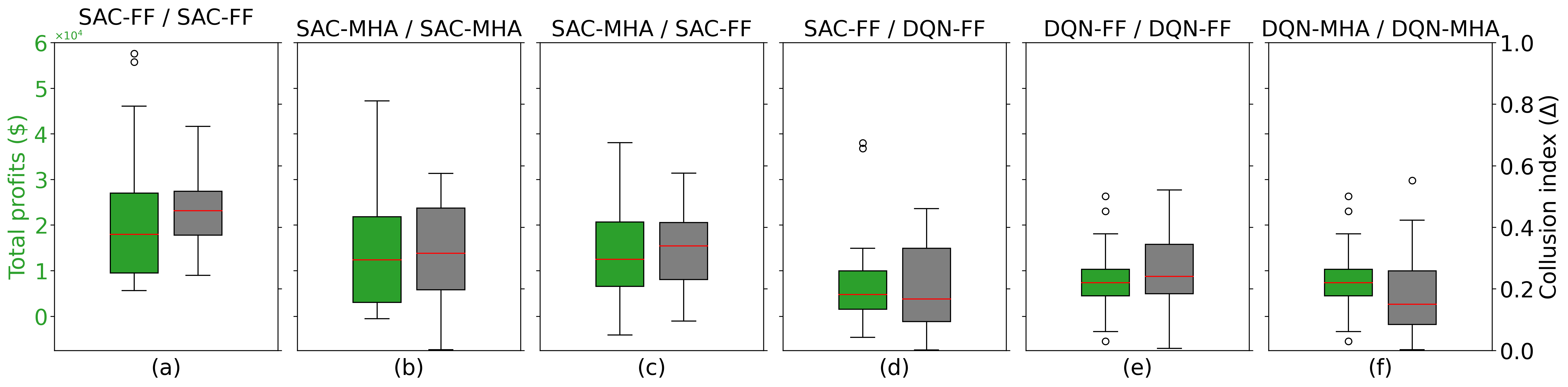}
    \caption{Box plots for total profit of the hubs and the resulting collusion index for 32 test scenarios}
    \label{fig: profit_curves}
\end{figure*}
We conduct an analysis of the sensitivity of the choice of the algorithm-architecture combination by the hubs on the pricing decisions and the resulting collusion index values. We consider five additional choice combinations by hub1 and hub2, which are SAC-MHA and SAC-MHA, SAC-MHA and SAC-FF, SAC-FF and DQN-FF, DQN-FF and DQN-FF, and DQN-MHA and DQN-MHA. The hourly average prices for all algorithm-architecture combinations are depicted in Figure \ref{fig: price_curves}. It can be observed that though the price profiles of the hubs for each combination have a similar pattern, the actual price values differ noticeably among the combinations. The prices across the hours are relatively higher when both hubs use the SAC algorithm. This also yields higher total daily profits. This is evident from the box plots for total profits in Figures \ref{fig: profit_curves}a, \ref{fig: profit_curves}b, and \ref{fig: profit_curves}c compared to those in Figures \ref{fig: profit_curves}d, \ref{fig: profit_curves}e, and \ref{fig: profit_curves}f. It can also be observed that the relative values of the hub prices compared to their costs of procuring power (which we refer to as markup) are higher in the first three combinations (see a, b, and c in Figure \ref{fig: price_curves}) with respect to the other combinations. As a result, the levels of tacit algorithmic collusion for these cases are also comparatively higher (see box plots for collusion index in Figure  \ref{fig: profit_curves}). The median values of the collusion index for all six combinations range from 0.18 to 0.45. This indicates that it is possible to have a low to moderate level of tacit algorithmic collusion among competing EV charging hubs. The level of collusion depends on the choice of algorithm-architecture combination that guides the pricing decisions.   

\begin{table}[htp]
\begin{tabular}{c|c|c|c|c|}
\cline{2-5}
                                                      & Test 1                                               & Test 2                                                 & Test 3                                                                      & Test 4                                                                      \\ \hline
\multicolumn{1}{|c|}{}                                & $H_0: \mu_{SAC} = \mu_{DQN}$ & $H_0: \mu^{MHA} = \mu^{FF}$    & $H_0: d\mu_{SAC}^{MHA} = d\mu_{SAC}^{FF}$ & $H_0: d\mu_{DQN}^{MHA} = d\mu_{DQN}^{FF}$ \\
\multicolumn{1}{|c|}{\multirow{-2}{*}{Hypothesis}}    & $H_1: \mu_{SAC} > \mu_{DQN}$ & $H_1: \mu^{MHA} \neq \mu^{FF}$ & $H_1: d\mu_{SAC}^{MHA} > d\mu_{SAC}^{FF}$ & $H_1: d\mu_{DQN}^{MHA} > d\mu_{DQN}^{FF}$ \\ \hline
\multicolumn{1}{|c|}{Sample size(n)} & 96 & 96 & 24 & 24 \\ \hline
\multicolumn{1}{|c|}{Test statisitc}                  &    5.78                                                  &     0.83                                                  &         2.54                                                                    &   5.05                                                                          \\ \hline
\multicolumn{1}{|c|}{Critical value}                  &         1.65                                             &      1.97                                                  &        1.66                                                                     &      1.66                                                                       \\ \hline
 
\multicolumn{1}{|c|}{p-value} &     1.44$e^{-4}$                                                 &                               0.40                         &             7$e^{-3}$                                                                &             3.67$e^{-6}$                                                                \\ \hline

\end{tabular}
\caption{Test of hypothesis for the impact of algorithm and architecture choices on the pricing outcomes}
\label{table: hypothesis test}
\end{table}
We observe from Figure \ref{fig: price_curves} that when the hubs adopt SAC, they appear to learn to price higher. This is perhaps because SAC promotes better exploration during training by sampling actions from the most recently learned action distributions while also maximizing entropy \cite{haarnoja2018soft}. On the other hand, the apparent lower prices resulting from the use of the DQN algorithm may be attributed to its inherently limited exploration strategy resulting from both action space discretizations and the exploration decay during learning. Regarding the choice of the neural network architecture, we observe that the choice of architecture (FF or MHA) does not seem to influence the average prices, whereas, the use of MHA architecture by both hubs yields hub prices that are further apart compared to those resulting from the use of FF architecture.  We conduct several hypothesis tests (with $\alpha = 0.05$) to validate the above observations, results of which are presented in Table \ref{table: hypothesis test}. Test 1 shows that the SAC algorithm yields higher average prices than the DQN algorithm, irrespective of the architecture choice. For this, we combined the data from Figure \ref{fig: price_curves}a and \ref{fig: price_curves}b for SAC and Figure \ref{fig: price_curves}e and \ref{fig: price_curves}f for DQN, resulting in a sample size of 96. Test 2 validates that the choice of architecture does not influence the average prices. For this we combine the data from Figure \ref{fig: price_curves}a and \ref{fig: price_curves}e for FF and Figure \ref{fig: price_curves}b and \ref{fig: price_curves}f for MHA, resulting in a sample size of 96. The outcomes of Tests 3 and 4 indicate that when both hubs use either SAC or DQN, MHA architecture yields higher differences among the hub prices ($d\mu$). Test 3 compares the hub price differences from Figure \ref{fig: price_curves}a with those from Figure \ref{fig: price_curves}b. Similarly, test 4 compares price differences from Figures \ref{fig: price_curves}e and \ref{fig: price_curves}f.

\begin{figure}[htp]
    \centering
\includegraphics[width=0.75\linewidth]{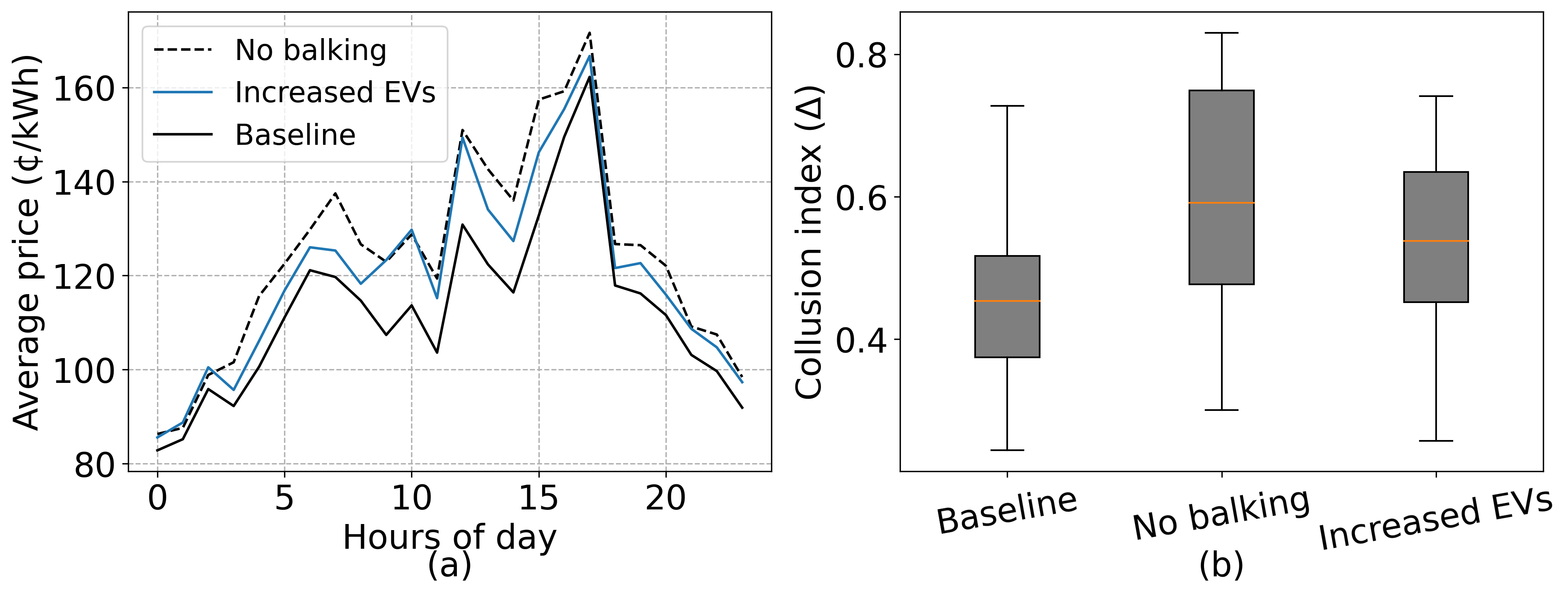}
    \caption{Impact of balking on pricing decisions and collusion index}
    \label{fig: sensitivity_balking_demand}
\end{figure}

Balking by EV owners modifies the effective EV arrival rate and the resulting charging demand for the hubs. We examine the impact of balking on the pricing decisions and the collusion level. We use the combination SAC-FF for both hubs and the balking probabilities used in our numerical case study as our baseline.
We considered two variants of the baseline case: 1) balking probabilities reduced to zero and 2) balking probabilities maintained as in baseline with a 50\% increase in the number of EVs seeking to charge. The results are shown in Figure \ref{fig: sensitivity_balking_demand}. When balking is removed, the EV owners always charge when a hub is available irrespective of prices. This keeps the demand intact and hence the hubs learn to set higher prices, which results in higher collusion. When the number of EVs seeking to charge is increased by 50\% with the balking probabilities maintained at baseline values, the effective charging demand was still lower than in the variant with no balking. Hence, the price and collusion level for the variant fell in between those for baseline and no balking. In essence, the price sensitivity of the EV owners (via balking) modulates the effective demand, which in turn influences the pricing decisions and the resulting collusion index.  
  
\section{Concluding remarks}\label{sec: conclusions}
As the competitors in pricing games are increasingly adopting artificial intelligence (AI) based learning algorithms for pricing decisions, it is raising the potential for tacit algorithmic collusion among the players who otherwise act independently. This is a critical concern for the antitrust agencies (e.g., the Federal Trust Commission (FTC) in the U.S.) responsible for limiting collusion and keeping marketplaces fair for consumers. In recent years, researchers have begun examining the extent of possible tacit algorithmic collusion in some limited versions of pricing games in canonical forms. These studies present contrasting findings ranging from no collusion to high collusion. In this paper, we have approached the question using price competition among EV charging hubs.  We do so by developing a two-step methodology for the hub-pricing game using the combination of a mixed integer linear program and a multi-agent DRL approach. The methodology is implemented on a sample numerical problem to demonstrate its ability to derive hub pricing strategies with their associated profits and algorithmic collusion levels. 

Since there are many choices for both the DRL algorithm and the neural network architecture that a hub can adopt to develop its pricing strategies, we examine six such algorithm-architecture combinations. Results from our numerical case study show that DRL-guided competing hubs can settle on charge prices that are on average 1.26 times higher than the cost of electric power. It is also shown that the pricing strategies yield collusion index values ranging from 0.18 to 0.45, which increases to 0.6 when balking is not considered, i.e., all arriving EVs charge irrespective of the prices. Following the characterization used in the existing literature, we conclude from our results that there can be a low to moderate level of tacit algorithmic collusion in DRL-guided pricing games with characteristics of Bertrand-Edgeworth games. Furthermore, a statistical comparison reveals that the choice of DRL algorithms by the players has a significant impact on the prices and the resulting collusion levels.

The scope of profitability through strategic pricing, as elucidated by our work, will serve to draw new entrants to the fast-charging business. Our model also provides a useful tool for antitrust agencies to quantify the level of tacit algorithmic collusion and decide if any regulations are needed to contain it. Following are the main limitations of our work, which primarily arise from how our methodology is implemented. We use a coarse time interval, an hour, to limit computational needs while RT market prices can change every 10-15 minutes interval. Our consideration of two identical hubs with the same parameters for the number of charging stations, DA commitment, and BSS characteristics is a simplification of the case study intended to allow better explainability of the results. Finally, we consider that the competing hubs update their neural network weights synchronously. In reality, since the hubs do not communicate, updating of the networks will most likely be asynchronous.

\appendices\label{appendix}
\section{Day-ahead commitment model}\label{Appendix: DA commitment model}
Each hub $i \in \mathcal{I}$ solves the following scenario-based stochastic DA commitment model. The objective function \eqref{eq: da com obj} maximizes the expected gross profit over all scenarios $(\omega \in \Omega$) by considering the charging revenue and the costs of DA, RT, and BSS powers. 
\begin{equation}\label{eq: da com obj}
\begin{aligned}
   &\max_{da_t^{i,com}} \sum_{\omega \in \Omega} \pi_{w} \sum_{ t \in T} \Bigg[(\widehat{p}_{\omega,t}^{i,ev} - p_{\omega,t}^{da}) da_{\omega,t}^{i,ev}
        + (\widehat{p}_{\omega,t}^{i,ev} bss_{t}^{i,ev})\\
        &+(\widehat{p}_{\omega,t}^{i,ev} - p_{\omega,t}^{rt}) rt_{\omega,t}^{i,ev}
        + \big(\min(p_{\omega,t}^{da},p_{\omega,t}^{rt} )- p_{\omega,t}^{da}\big) da_{\omega,t}^{i,rt} 
        \\& + (- p_{\omega, t}^{i,da}) da_{t}^{i,bss}\Bigg] 
\end{aligned}
\end{equation}
    subject to: 
    \begin{equation}\label{eq: da com da balance}
        da_{\omega,t}^{i,ev} + da_{\omega,t}^{i,bss} + da_{\omega,t}^{i,rt} = da_{t}^{i,com},\quad \forall \omega \in \Omega, \forall t \in T.
    \end{equation}
    \begin{equation}\label{eq: da com load balance}
        da_{\omega,t}^{i,ev} + bss_{\omega,t}^{i,ev} + rt_{\omega,t}^{i,ev} = \widehat{ev}_{\omega,t}^{i,load}, \quad \forall \omega \in \Omega, \forall t \in T.
    \end{equation}
    \begin{equation}\label{eq: da com bss balance}
        \phi_{\omega, t}^{i} = \phi_{\omega, t-1}^{i} + da_{\omega,t}^{i,bss} - bss_{\omega,t}^{i,ev}, \quad \forall \omega \in \Omega, \forall t \in T.
    \end{equation}
    \begin{equation}\label{eq: da com da_ev}
        da_{\omega, t}^{i,ev} \geq \Psi \widehat{ev}_{\omega,t}^{i,load}, \quad \forall \omega \in \Omega, \forall t \in T.
    \end{equation}
    \begin{equation}\label{eq: da com charge discharge}
        x_{\omega, t}^i + y_{\omega, t}^i \leq 1, \quad \forall \omega \in \Omega, \forall t \in T.
    \end{equation}
    \begin{equation}\label{eq: da com bss ch ub 1}
        bss_{\omega,t}^{i,ch} \leq M_{rate}^{i,ch} x_{\omega,t}^i, \quad \forall \omega \in \Omega, \forall t \in T.
    \end{equation}
    \begin{equation}\label{eq: da com bss ch ub 2}
        bss_{\omega,t}^{i,ch} \leq M_\phi^i - \phi_{\omega,t}^i, \quad \forall \omega \in \Omega, \forall t \in T.
    \end{equation}
    \begin{equation}\label{eq: da com bss dch ub 1}
        bss_{\omega,t}^{i,dch} \leq M_{rate}^{i,dch} y_{\omega,t}^i, \quad \forall \omega \in \Omega, \forall t \in T.
    \end{equation}
    \begin{equation}\label{eq: da com bss dch ub 2}
        bss_{\omega,t}^{i,ch} \leq \phi_{\omega,t}^i - m_\phi^i, \quad \forall \omega \in \Omega, \forall t \in T.
    \end{equation}
    \begin{equation}\label{eq: da com non negativity} da_{\omega,t}^{i,ev}, bss_{\omega,t}^{i,ev}, rt_{\omega,t}^{i,ev}, da_{\omega,t}^{i,bss},
    da_{\omega,t}^{i,rt}
    \geq 0, \quad \forall \omega \in \Omega, \forall t \in T.
    \end{equation}
    \begin{equation}\label{eq: binary}
        x_{\omega, t}^{i}, y_{\omega, t}^i \in \{0, 1\}, \quad \forall \omega \in \Omega, \forall t \in T.
    \end{equation}

Let $\pi_\omega$ represent the probability associated with scenario $\omega \in \Omega$. Since the DA commitment is made prior to the pricing decisions, the minimum of DA and RT prices
($\widehat{p}_{\omega,t}^{i,ev}$) is used as the surrogate for the EV charging price in the DA model. Let $da_{\omega,t}^{ev}$, $da_{\omega,t}^{rt}$, and $da_{\omega,t}^{bss}$ be the DA power used for charging EVs, sending back to the grid, and charging the BSS, respectively. Similarly, let $rt_{\omega, t}^{rt}$ be the RT power used to charge the EVs. The first three elements of the objective function are profits from charging EVs using DA $(da_{\omega, t}^{i,ev})$, BSS $(bss_{\omega, t}^{i,ev})$, and RT power $(rt_{\omega, t}^{i,ev})$, respectively. The fourth element considers the loss, if any, for sending unused DA commitments to the RT market $(da_{\omega, t}^{i,rt})$. The fifth element considers the cost of charging BSS. It is assumed that the BSS is charged only using DA power and BSS is discharged only for charging EVs. The constraints address the following: power balance for DA commitment $(da_{\omega, t}^{i,com})$, aggregated EV charging demand $(\hat{ev}_{\omega, t}^{i,load})$, and the BSS $(\phi_{\omega, t}^{i})$  in \eqref{eq: da com da balance}, \eqref{eq: da com load balance}, and \eqref{eq: da com bss balance}, respectively; minimum charging power need that must be committed to the DA market \eqref{eq: da com da_ev};  the BSS can either charge $(x_{\omega,t}^{i} = 1 \& y_{\omega,t}^{i} = 0) $ or discharge $(x_{\omega,t}^{i} = 0 \& y_{\omega,t}^{i} = 1)$ or kept idle $(x_{\omega,t}^{i} = 0 \& y_{\omega,t}^{i} = 0)$ \eqref{eq: da com charge discharge}; the upper limits of the amount of charge the BSS can accept, \eqref{eq: da com bss ch ub 1} and \eqref{eq: da com bss ch ub 2}; the upper limits of the amount of BSS discharge, \eqref{eq: da com bss dch ub 1} and \eqref{eq: da com bss dch ub 2} with $M_{rate}^{i, ch}$ being the maximum charge rate and $M_\phi^{i}$ being the maximum storage capacity; and the nonnegativity of decision variables \eqref{eq: da com non negativity}.  Constraint \eqref{eq: binary} represents the binary nature of variables $x_{\omega,t}^{i}$ and $y_{\omega,t}^{i}$.

\section{Power management model}\label{Appendix: Power mgmt}
Each hub $i \in \mathcal{I}$ solves the following power management model for each time period $t \in T$ after the EVs make their hub selection. 
\begin{equation}\label{eq: pmgt obj}
\begin{aligned}
        &\max  \big[(p_{t}^{i,ev} - p_{t}^{da}) da_{t}^{i,ev} + (p_{t}^{i,ev} - p_{t}^{i,bss}) bss_{t}^{i,ev} 
        \\&+ (p_{t}^{i,ev} - p_{t}^{rt}) rt_{t}^{i,ev}  + \big(\min(p_{t}^{da}, p_{t}^{rt}) - p_{t}^{da}\big) da_{t}^{i,rt}\big]
\end{aligned}
    \end{equation}
    subject to: 
    \begin{equation}\label{eq: pmgt da ev}
        da_t^{i,ev} = \min \Big( ev_t^{i,load}, da_t^{i,com} \Big).
    \end{equation}
    \begin{equation}\label{eq: pmgt da balance}
        da_{t}^{i,ev} + da_{t}^{i,bss} + da_{t}^{i,rt} = da_{t}^{i,com}.
    \end{equation}
    \begin{equation}\label{eq: pmgt ev load balance}
        da_{t}^{i,ev} + bss_{t}^{i,ev} + rt_{t}^{i,ev} = ev_{t}^{i,load}.
    \end{equation}
    \begin{equation}\label{eq: pmgt bss balance}
        \phi_{t}^i = \phi_{t-1}^i + da_t^{i,bss} - bss_t^{i,ev}.
    \end{equation}
    \begin{equation}\label{eq: pmgt bss charge discharge}
        x_t^i + y_t^i \leq 1.
    \end{equation}
    \begin{equation}\label{eq: pmgt bss charge ub1}
        da_{t}^{i,bss} \leq M_{rate}^{i,ch} x^i.
    \end{equation}
    \begin{equation}\label{eq: pmgt bss charge ub2}
        da_{t}^{i,bss} \leq M_\phi^i - \phi_t^i.
    \end{equation}
    \begin{equation}\label{eq: pmgt bss discharge ub1}
        bss_{t}^{i,ev} \leq M_{rate}^{i,dch} y^i.
    \end{equation}
    \begin{equation}\label{eq: pmgt bss discharge ub2}
        bss_{t}^{i,ev} \leq  \phi_t^i - m_\phi^i.
    \end{equation}
    \begin{equation}\label{eq: pmgmt non negativity}
        da_{t}^{i,ev}, bss_{t}^{i,ev}, rt_{t}^{i,ev}, da_{t}^{i,bss}, bss_{t}^{i,ev} \geq 0.
    \end{equation}
The objective function \eqref{eq: pmgt obj} maximizes the gross profit. Constraint \eqref{eq: pmgt da ev} ensures that the DA commitment is prioritized for EV charging. Power balance equations for DA commitment, EV charging, and the BSS are accounted for in \eqref{eq: pmgt da balance}, \eqref{eq: pmgt ev load balance}, and \eqref{eq: pmgt bss balance}, respectively. The BSS can only be either charged or discharged or remain idle at any time period \eqref{eq: pmgt bss charge discharge}. Constraints \eqref{eq: pmgt bss charge ub1} and \eqref{eq: pmgt bss charge ub2} maintain the maximum charging limits. Similarly, constraints \eqref{eq: pmgt bss discharge ub1} and \eqref{eq: pmgt bss discharge ub2} maintain the maximum discharging limits. Constraint \eqref{eq: pmgmt non negativity} maintains the non-negativity of the power management decision variables.

\section{Overview of DRL algorithms}\label{Appendix: DRL algo}
\subsection{Deep-Q network (DQN)}

DQN is a form of Q learning that employs neural networks to estimate the Q values of state-action pairs \cite{mnih2013playing}. The goal of DQN is to minimize the squared error between the predicted Q-network, characterized by parameters $\theta$, and the estimated actual Q values provided by the target network with parameters $\theta^-$, as shown in \eqref{eq: DQN obj}.

\begin{equation}\label{eq: DQN obj}
    \mathcal{L}(\theta) = \mathbb{E}_{s,a,r,s' \sim \mathcal{D}} \Bigg[ \Big( Q(s,a; \theta\Big) - (r + \gamma \max_{a'}Q(s',a';\theta^-)\Big)^2\Bigg].
\end{equation}
The weight of the Q network is updated by calculating the gradient of the loss as in \eqref{eq: DQN gradient update}.

\begin{equation}\label{eq: DQN gradient update}
    \theta = \theta - \frac{\delta\mathcal{L}}{\delta\theta}.
\end{equation}
The target network ($\theta^-$) is updated once every C number of Q network updates \eqref{eq: DQN target update}. 

\begin{equation}\label{eq: DQN target update}
    \theta^- = \tau \theta^- + (1 - \tau) \theta,
\end{equation}
where $\tau$ is a weight with a value ranging from 0 to 1.

\subsection{Soft-actor critic (SAC)}

The SAC belongs to the category of policy gradient algorithms based on the actor-critic framework \cite{haarnoja2018soft}. In this setup, the actor-network learns the policy, mapping states to actions, while the critic evaluates the state-action pair's value and aids the actor through feedback. The actor-network $\pi$, characterized by parameter $\phi$, aims to maximize both the reward $r$ and the entropy $\mathcal{H}$ shown in \eqref{eq: sac actor obj}. 


\begin{equation}\label{eq: sac actor obj}
    \mathcal{J}_\pi(\phi) = \mathbb{E}_{(s,a) \sim \pi_\phi} \Bigg[r(s,a) + \alpha\mathcal{H}(\pi(\cdot|s)) \Bigg].
\end{equation}
The objective of the critic network Q, represented by parameter $\theta$, is to minimize the squared error of Q-values which is represented in \eqref{eq: sac critic obj}.
\begin{equation}\label{eq: sac critic obj}
    \mathcal{J}_Q(\theta) = \mathbb{E}_{(s,a) \sim \mathcal{D}}\Bigg[ \frac{1}{2}\Big(Q_\theta(s,a) - \hat{Q}_{
    \hat{\theta}}(s,a) \Big)^2\Bigg].
\end{equation}
In the above equation, $\hat{Q}$ is the target critic network represented by parameters $\hat{\theta}$. The actor/critic network weights as well as the target network weights are updated as in \eqref{eq: DQN gradient update} and \eqref{eq: DQN target update}, respectively. 

\section{Schematic representations of neural network architectures}\label{Appendix: network}

\begin{figure}[htp]
    \centering
    \includegraphics[width=1\linewidth]{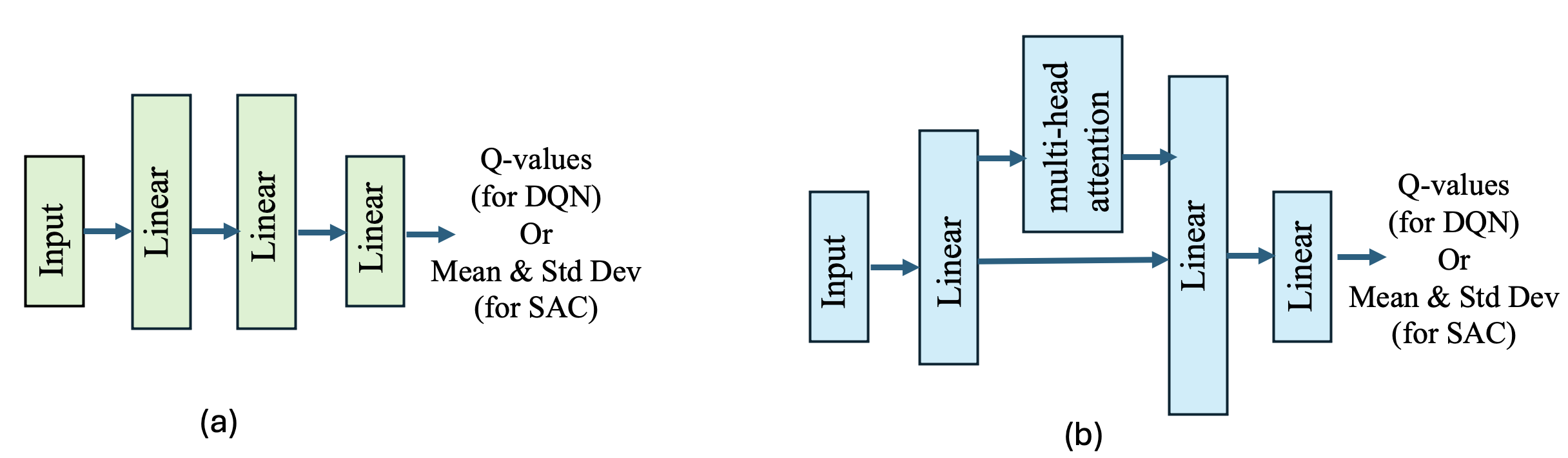}
    \caption{Schematic representations of a) FF and b) MHA network architecture}
    \label{fig: network architectures}
\end{figure}

\begin{figure}[htp]
    \centering
    \includegraphics[width=0.35\linewidth]{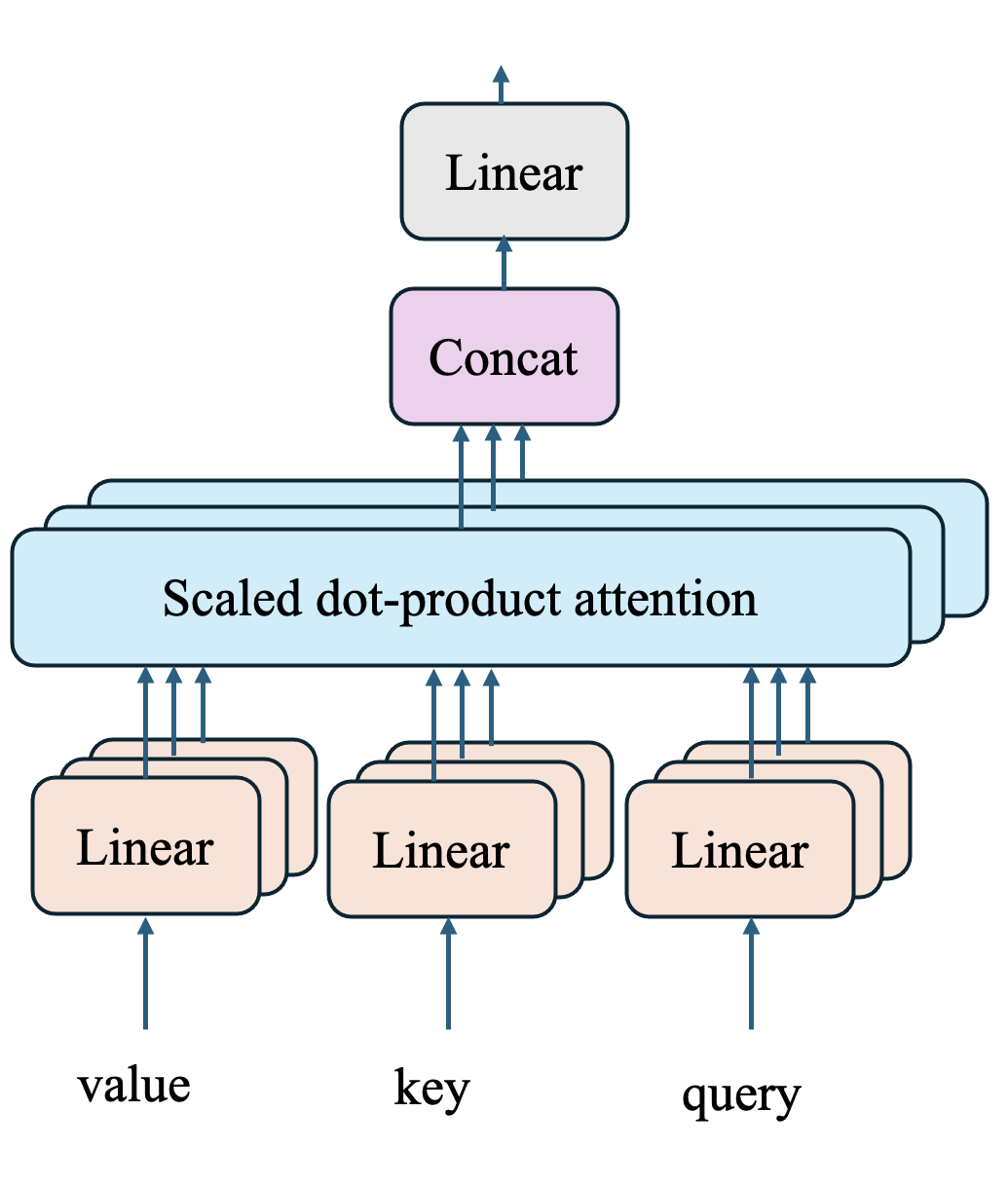}
    \caption{Schematic representations of three-head attention layer \cite{vaswani2017attention}}
    \label{fig: mha}
\end{figure}

Figures \ref{fig: network architectures}a and \ref{fig: network architectures}b depict the schematic representations of feed-forward (FF) and multi-head attention (MHA) network architectures, respectively. The FF architecture comprises linear layers exclusively, while the MHA architecture incorporates an additional three-head attention layer (as shown in Figure \ref{fig: mha}). Both of these architectures, when applied in the context of DQN, produce Q-values for all actions given a particular state. For DQN, the action with the highest Q-value is typically chosen with a high probability, while there remains a small chance of selecting any other feasible action. When used as SAC actor network, these architectures yield the mean and standard deviation for the action, which is then sampled from a Normal distribution. The corresponding SAC critic networks mirror the structure of the actor networks, with the inclusion of the action in the input layer and the final layer outputting the Q-value for the specific state-action pair. 

\section{Hyperparameters}\label{Appendix: Hyperparameters}

The hyperparameters and corresponding values associated with the neural network architectures and the DRL algorithms are provided in Table \ref{table: hyperparameters}. Note: The action choices for the DQN algorithm include 101 possible values $\{1.0, 1.01, \dots, 1.99, 2.0\}$. The selected action choice is multiplied with $\min(p_t^{da}, p_t^{rt})$ to find the actual pricing decision. 
\begin{table}[htp]
\centering
\begin{tabular}{llll}
\toprule
Parameters & Value\\
\midrule
 Optimizer & Adam \\
 Actor learning rate  & 3 $\times$ $10^{-4}$  \\
 Critic earning rate &  3 $\times$ $10^{-4}$ \\
 Discount factor ($\gamma$)  & 0.99  \\
 Replay size& $10^6$ \\
 Number of hidden layers & 2 \\
 Number of units per hidden layer & 265 \\
 Entropy regularization coefficient  & 0.2 \\ 
 Number of samples per minibatch & 256 \\
 Activation function & RelU\\
Target smoothing coefficient ($\tau$) & 0.05\\
 Target update interval & 2\\
 Initialization & Xavier uniform\\
 Input dimension  & 7\\
 Action dimension (SAC) & 1\\
 Action dimension (DQN) & 100\\
Target entropy & -1\\
Number of heads in MHA  & 2 \\
 Epsilon start (DQN)  & 0.8\\
 Epsilon end (DQN) & 0.05\\ 
Epsilon decay frequency (DQN) & 40 $\times$ $10^{3}$\\
 \bottomrule
 \end{tabular}
  \caption{Hyperparameters for architectures and algorithms}
 \label{table: hyperparameters}
\end{table}

\newpage
\bibliographystyle{unsrt}

\bibliography{references}

\begin{thebibliography}{10}

\bibitem{calvano2020artificial}
Emilio Calvano, Giacomo Calzolari, Vincenzo Denicolo, and Sergio Pastorello.
\newblock Artificial intelligence, algorithmic pricing, and collusion.
\newblock {\em American Economic Review}, 110(10):3267--3297, 2020.

\bibitem{mellgren2020tacit}
Filip Mellgren.
\newblock Tacit collusion with deep multi-agent reinforcement learning, 2020.

\bibitem{den2022artificial}
Arnoud~V den Boer, Janusz~M Meylahn, and Maarten~Pieter Schinkel.
\newblock Artificial collusion: Examining supracompetitive pricing by q-learning algorithms.
\newblock {\em Amsterdam Law School Research Paper}, (2022-25), 2022.

\bibitem{zhang2023pricing}
Weipeng Zhang.
\newblock Pricing via artificial intelligence: The impact of neural network architecture on algorithmic collusion.
\newblock {\em Available at SSRN 4489688}, 2023.

\bibitem{paudel2022infrastructure}
Diwas Paudel and Tapas~K Das.
\newblock {Infrastructure planning for ride-hailing services using shared autonomous electric vehicles}.
\newblock {\em International Journal of Sustainable Transportation}, pages 1--16, 2023.

\bibitem{alterenative_fuel}
{Alternative Fuels Data Center}.
\newblock Electric vehicle charging infrastructure trends.
\newblock \url{https://afdc.energy.gov/fuels/electricity_infrastructure_trends.html}, 2024.
\newblock Accessed: 2024-02-12.

\bibitem{paudel2023distributionally}
Diwas Paudel, Nicolas Bustos, and Tapas~K Das.
\newblock A distributionally robust approach for day-ahead power procurement by ev charging hubs.
\newblock In {\em 2023 IEEE Power \& Energy Society General Meeting (PESGM)}, pages 1--5. IEEE, 2023.

\bibitem{yan2018optimized}
Qin Yan, Bei Zhang, and Mladen Kezunovic.
\newblock Optimized operational cost reduction for an ev charging station integrated with battery energy storage and pv generation.
\newblock {\em IEEE Transactions on Smart Grid}, 10(2):2096--2106, 2018.

\bibitem{paudel2023deep}
Diwas Paudel and Tapas~K Das.
\newblock A deep reinforcement learning approach for power management of battery-assisted fast-charging ev hubs participating in day-ahead and real-time electricity markets.
\newblock {\em Energy}, 283:129097, 2023.

\bibitem{hicks1935annual}
John~R Hicks.
\newblock Annual survey of economic theory: The theory of monopoly.
\newblock {\em Econometrica: Journal of the Econometric Society}, pages 1--20, 1935.

\bibitem{dixon1984existence}
Huw Dixon.
\newblock The existence of mixed-strategy equilibria in a price-setting oligopoly with convex costs.
\newblock {\em Economics Letters}, 16(3-4):205--212, 1984.

\bibitem{burman2021deep}
Vibhati Burman, Rajesh~Kumar Vashishtha, Rajan Kumar, and Sharadha Ramanan.
\newblock {Deep reinforcement learning for dynamic pricing of perishable products}.
\newblock In {\em Optimization and Learning: 4th International Conference, OLA 2021, Catania, Italy, June 21-23, 2021, Proceedings 4}, pages 132--143. {Springer}, 2021.

\bibitem{kastius2021dynamic}
Alexander Kastius and Rainer Schlosser.
\newblock Dynamic pricing under competition using reinforcement learning.
\newblock {\em Journal of Revenue and Pricing Management}, pages 1--14, 2021.

\bibitem{aljafari2023electric}
Belqasem Aljafari, Pandia~Rajan Jeyaraj, Aravind~Chellachi Kathiresan, and Sudhakar~Babu Thanikanti.
\newblock Electric vehicle optimum charging-discharging scheduling with dynamic pricing employing multi agent deep neural network.
\newblock {\em Computers and Electrical Engineering}, 105:108555, 2023.

\bibitem{fang2020dynamic}
Cheng Fang, Haibing Lu, Yuan Hong, Shan Liu, and Jasmine Chang.
\newblock Dynamic pricing for electric vehicle extreme fast charging.
\newblock {\em IEEE Transactions on Intelligent Transportation Systems}, 22(1):531--541, 2020.

\bibitem{narayan2022dynamic}
Ajay Narayan, Aakash Krishna, Prasant Misra, Arunchandar Vasan, and Venkatesh Sarangan.
\newblock A dynamic pricing system for electric vehicle charging management using reinforcement learning.
\newblock {\em IEEE Intelligent Transportation Systems Magazine}, 14(6):122--134, 2022.

\bibitem{abdalrahman2020dynamic}
Ahmed Abdalrahman and Weihua Zhuang.
\newblock Dynamic pricing for differentiated pev charging services using deep reinforcement learning.
\newblock {\em IEEE Transactions on Intelligent Transportation Systems}, 23(2):1415--1427, 2020.

\bibitem{liu2021dynamic}
Dunnan Liu, Weiye Wang, Lingxiang Wang, Heping Jia, and Mengshu Shi.
\newblock Dynamic pricing strategy of electric vehicle aggregators based on ddpg reinforcement learning algorithm.
\newblock {\em IEEE access}, 9:21556--21566, 2021.

\bibitem{deng2018demand}
Qian Deng, Sujit Tripathy, Daniel Tylavsky, Travis Stowers, and Jeff Loehr.
\newblock Demand modeling of a dc fast charging station.
\newblock In {\em 2018 North American Power Symposium (NAPS)}, pages 1--6. IEEE, 2018.

\bibitem{Idaho}
{Idaho National Laboratory}.
\newblock Plug-in electric vehicle and infrastructure analysis.
\newblock \url{https://inldigitallibrary.inl.gov/sites/sti/sti/6799570.pdf}, 2015.

\bibitem{funke2020fast}
Simon Funke, Patrick Jochem, Sabrina Ried, and Till Gnann.
\newblock Fast charging stations with stationary batteries: A techno-economic comparison of fast charging along highways and in cities.
\newblock {\em Transportation Research Procedia}, 48:3832--3849, 2020.

\bibitem{hussain2020optimal}
Akhtar Hussain, Van-Hai Bui, and Hak-Man Kim.
\newblock Optimal sizing of battery energy storage system in a fast ev charging station considering power outages.
\newblock {\em IEEE Transactions on Transportation Electrification}, 6(2):453--463, 2020.

\bibitem{traffic_flow}
FDOT.
\newblock Traffic information.
\newblock \url{https://www.fdot.gov/statistics/trafficdata}, 2023.
\newblock (accessed : 03 March 2023).

\bibitem{McKinsey}
McKinsey and Company.
\newblock Why the automotive future is electric.
\newblock \url{https://www.mckinsey.com/industries/automotive-and-assembly/our-insights/why-the-automotive-future-is-electric}, 2021.

\bibitem{mnih2013playing}
Volodymyr Mnih, Koray Kavukcuoglu, David Silver, Alex Graves, Ioannis Antonoglou, Daan Wierstra, and Martin Riedmiller.
\newblock Playing atari with deep reinforcement learning.
\newblock {\em arXiv preprint arXiv:1312.5602}, 2013.

\bibitem{haarnoja2018soft}
Tuomas Haarnoja, Aurick Zhou, Kristian Hartikainen, George Tucker, Sehoon Ha, Jie Tan, Vikash Kumar, Henry Zhu, Abhishek Gupta, Pieter Abbeel, et~al.
\newblock Soft actor-critic algorithms and applications.
\newblock {\em arXiv preprint arXiv:1812.05905}, 2018.

\bibitem{vaswani2017attention}
Ashish Vaswani, Noam Shazeer, Niki Parmar, Jakob Uszkoreit, Llion Jones, Aidan~N Gomez, {\L}ukasz Kaiser, and Illia Polosukhin.
\newblock Attention is all you need.
\newblock {\em Advances in neural information processing systems}, 30, 2017.

\bibitem{PJM}
PJM.
\newblock Energy market.
\newblock \url{https://www.pjm.com/markets-and-operations/energy.aspx}, 2022.
\newblock (accessed : 2 July 2022).

\bibitem{green2014divide}
Richard Green, Iain Staffell, and Nicholas Vasilakos.
\newblock Divide and conquer? k-means clustering of demand data allows rapid and accurate simulations of the british electricity system.
\newblock {\em IEEE Transactions on Engineering Management}, 61(2):251--260, 2014.

\bibitem{schwalbe2018algorithms}
Ulrich Schwalbe.
\newblock Algorithms, machine learning, and collusion.
\newblock {\em Journal of Competition Law \& Economics}, 14(4):568--607, 2018.

\bibitem{ittoo2017algorithmic}
Ashwin Ittoo and Nicolas Petit.
\newblock Algorithmic pricing agents and tacit collusion: A technological perspective.
\newblock {\em Chapter in L'intelligence artificielle et le droit, Herv{\'e} JACQUEMIN and Alexandre DE STREEL (eds), Bruxelles: Larcier}, pages 241--256, 2017.

\end{thebibliography}

%

\end{document}